\PassOptionsToPackage{compress}{natbib}

\documentclass{article}

\usepackage[preprint]{neurips_2026}

\usepackage[utf8]{inputenc} 
\usepackage[T1]{fontenc}    
\usepackage{hyperref}       
\usepackage{url}            
\usepackage{booktabs}       
\usepackage{amsfonts}       
\usepackage{nicefrac}       
\usepackage{microtype}      
\usepackage{xcolor}         
\usepackage{amsmath}
\usepackage[pdftex]{graphicx}
\usepackage{wrapfig}
\usepackage{multirow}
\usepackage{makecell}
\usepackage[table]{xcolor}
\usepackage{soul} 
\usepackage{algorithm}
\usepackage{algpseudocode}
\usepackage{caption}
\usepackage[most]{tcolorbox}
\usepackage{fvextra}
\usepackage{upquote}
\usepackage{tabularx}
\usepackage{array}

\providecommand{\best}[1]{\textcolor{red}{\textbf{#1}}}
\providecommand{\second}[1]{\textcolor{red}{\underline{#1}}}
\providecommand{\blackcell}[1]{\cellcolor{blackrowgray}#1}
\providecommand{\third}[1]{\textcolor{red}{#1}}

\definecolor{blackrowgray}{gray}{0.94}
\DefineVerbatimEnvironment{PromptVerbatim}{Verbatim}{
  fontsize=\scriptsize,
  breaklines=true,
  breakanywhere=true,
  breaksymbolleft={},
  breaksymbolright={},
  breakindent=0pt,
  tabsize=2
}
\newtcolorbox{promptbox}[1][]{
  colback=gray!3,
  colframe=black!45,
  fonttitle=\bfseries,
  breakable,
  sharp corners,
  boxrule=0.5pt,
  left=6pt,
  right=6pt,
  top=6pt,
  bottom=6pt,
  #1
}
\title{Discovering Natural Transformation Vulnerabilities in Black-Box Vision Models}

\author{Dongsu Song \and
DaeYun GO \and
Jay Hoon Jung}

\begin{document}

\maketitle

\begin{abstract}
Natural adversarial examples (NAEs) reveal that vision models can fail under realistic semantic changes beyond norm-bounded perturbations. 
However, generating NAEs in a black-box setting remains challenging because existing generative attacks often rely on surrogate models, learned attack priors, or costly query-based optimization, whereas the natural transformations that expose model vulnerabilities are unknown a priori.
We propose \textbf{Adversarial Scenario Attack (ASA)}, a query-based black-box framework that searches over natural-language editing scenarios using a multimodal language model and a modern text-guided generative editor. ASA jointly explores background, weather, and material/color transformations through winner--loser feedback, and uses a greedy explorer to compose only attack-improving scenarios. 
Across diverse ImageNet classifiers, ASA achieves substantially higher attack success rates than prior query-based generative attacks while requiring fewer victim-model queries and preserving competitive perceptual quality.
Moreover, ASA exhibits both image-level and prompt-level transferability: its adversarial images remain effective across victim-model architectures, while its discovered editing scenarios can be reused across same-class images and, in some cases, across architectures.
These findings suggest that vision models possess reusable vulnerabilities to natural transformation patterns, which ASA can efficiently identify in a black-box setting.
\end{abstract}

\section{Introduction}
Deep learning models are known to be vulnerable to adversarial attacks, where carefully designed perturbations can induce incorrect predictions \cite{att_ref1, att_ref2, att_ref4}. Most prior studies focus on restricted attacks that constrain perturbations within an $\ell_p$ norm ball to preserve perceptual similarity \cite{att_ref3, pgd, ffsgm, rfpar}. However, robustness evaluation based only on small pixel-level perturbations may overlook realistic semantic variations. Recent advances in generative models have therefore motivated unrestricted adversarial attacks, which generate natural-looking modifications or Natural Adversarial Examples (NAEs) beyond explicit norm constraints \cite{advdiffuser, ACA, NCF, natadiff, NAE1, NAE2}.

Existing generative NAE methods remain limited for strict black-box attacks on a given input image. Many methods rely on white-box surrogate gradients or learned attack priors to guide the generation process, and then expect the resulting examples to transfer to black-box victim models \cite{advdiffuser, diffattack, ACA, ACGAN}. Other recent methods synthesize NAEs from random noise, improving naturalness and transferability but producing a new image of the same concept rather than adversarially editing the original input \cite{natadiff, ADVdiff}. 
Thus, natural adversarial editing of a given input image remains underexplored in query-limited black-box settings, particularly when neither victim-specific surrogate training nor attack-specific prior learning is available.

\begin{figure*}[t]
    \centering
    \includegraphics[width=\textwidth]{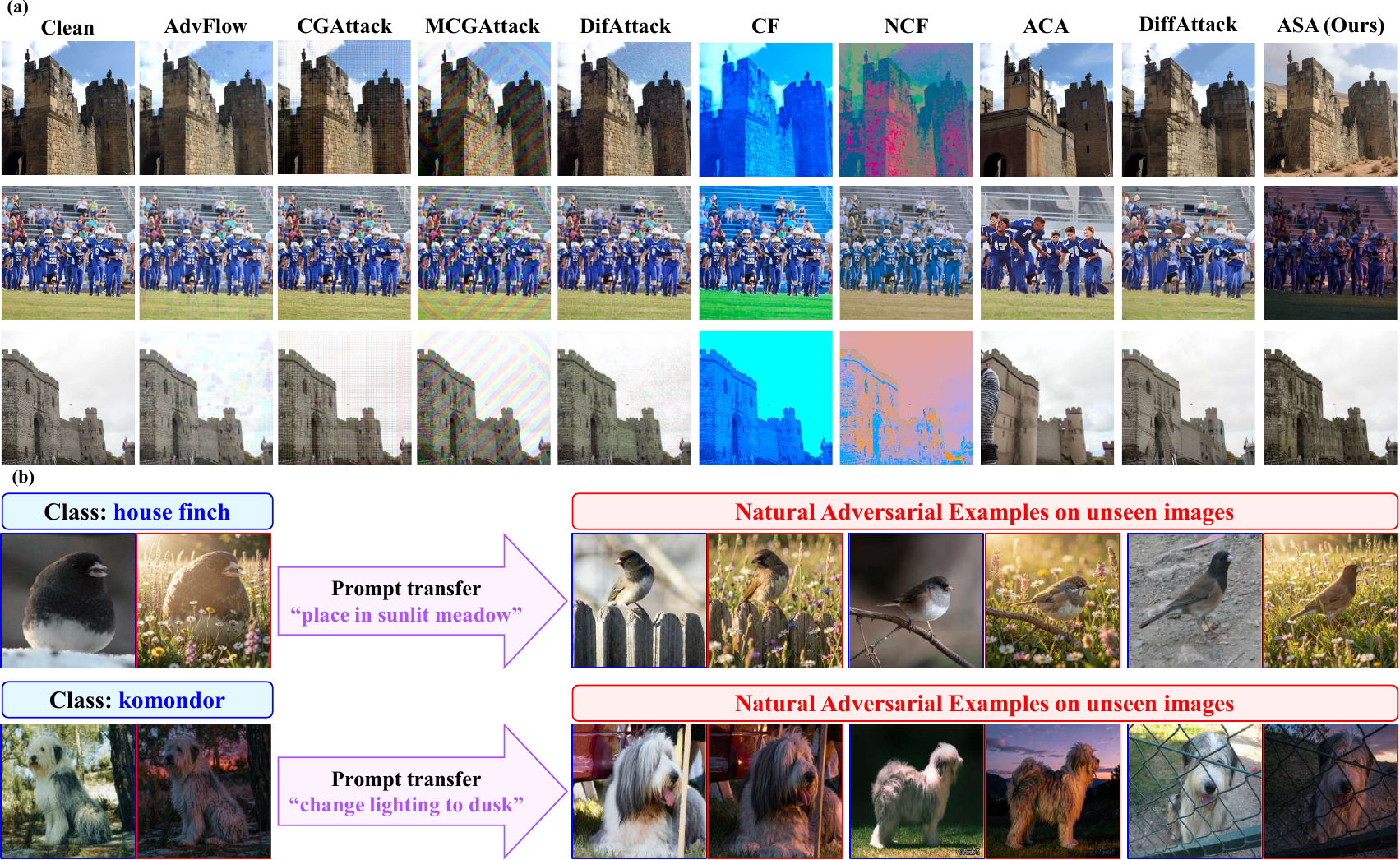}
    \vspace{-0.2cm}
    \caption{Qualitative overview of ASA. (a) ASA generates visually natural adversarial examples compared with existing attacks. (b) Same-class prompt transfer produces diverse NAEs using transferable prompts. Additional examples are provided in App.~\ref{app:vis}.}
    \label{fig:prompt_transferability}
    \vspace{-0.6cm}
\end{figure*}


Recent text-guided generative editors enable realistic, instruction-driven modifications to a given image, thereby expanding the adversarial search space from conventional pixel-level or latent-space perturbations to natural editing scenarios. However, existing query-based generative attacks have not fully exploited this scenario-level space, particularly under tight query budgets. To this end, we propose \textbf{Adversarial Scenario Attack (ASA)}, a query-based black-box framework that generates NAEs by directly searching over natural-language editing scenarios using FLUX.2 [klein] 9B-KV\cite{flux-2-klein}.


Prior studies on NAEs have shown that model failures can arise from realistic transformations in natural scenes, including changes in background \cite{advbackground}, weather \cite{advweather1, advweather2}, and surface appearance \cite{advtex1, advtex2}. However, for a given input, it is generally unclear a priori which type of natural transformation will expose the most vulnerable attributes of the victim model. Motivated by this observation, ASA jointly explores multiple natural editing scenarios rather than treating them as separate attack pipelines. Because this scenario space is open-ended and image-dependent, ASA uses a multimodal large language model (MLLM) \cite{gemma} to generate plausible strategy-specific transformation scenarios, which are then instantiated by a modern generative editor to produce natural edited candidates.
We further introduce a \textbf{Greedy Explorer}, which composes strategy-specific scenarios and retains only attack-score-improving combinations, enabling ASA to accumulate complementary natural transformations. As a result, ASA generates effective black-box NAEs with improved query efficiency, without task-specific surrogate training or attack-prior learning, while also showing competitive image-level transferability and promising prompt-level transferability. 
Our contributions are summarized as follows:
\begin{itemize}
\item \textbf{Query-efficient black-box NAE generation:} We propose
\textbf{Adversarial Scenario Attack (ASA)}, a black-box method that searches and
composes natural-language editing scenarios to produce NAEs without attack-specific preparation. ASA achieves higher
attack success with fewer victim-model queries than prior generative attacks.

\item \textbf{Competitive transferability and visual similarity:} ASA generates
adversarial images that preserve visual similarity while achieving
transferability competitive with existing transfer-based attacks, despite no
access to model internals.

\item \textbf{Prompt-level transferability:} We show that discovered adversarial prompts often transfer across images of the same class and sometimes across architectures, suggesting reusable natural transformation vulnerabilities.
\end{itemize}

\section{Related work}
\subsection{Generative models for natural adversarial examples}
\begin{figure*}[t]
    \centering
    \includegraphics[width=\textwidth]{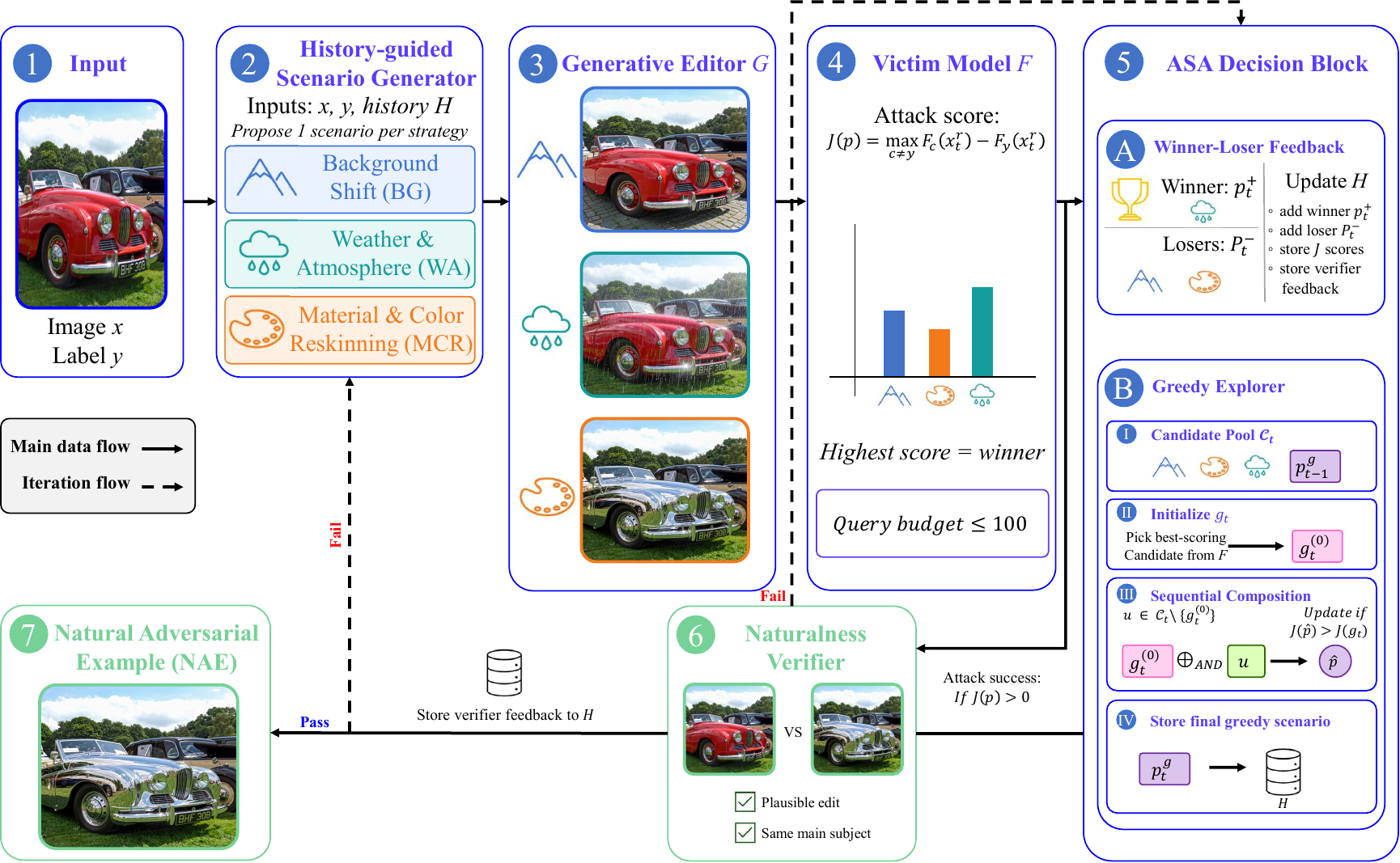}
    \vspace{-0.2cm}
    \caption{
    Overview of Adversarial Scenario Attack (ASA). The method searches over natural-language editing scenarios by combining winner--loser feedback with a Greedy Explorer that composes promising candidates into a stronger scenario, followed by naturalness verification.
    }
    \label{fig:overview}
    \vspace{-0.6cm}
\end{figure*}
Generative and semantic image-editing approaches have been widely explored for
constructing NAEs. Early work synthesized unrestricted adversarial examples from
scratch using an AC-GAN-based approach and latent-space search \cite{ACGAN}.
Other studies investigated semantic-level manipulations that preserve the
overall structure of an image while altering attributes such as color, texture,
or contextual appearance. Representative examples include \textit{ColorFool
(CF)} \cite{CF} and \textit{Natural Color Fool (NCF)} \cite{NCF}, which use
semantic color transformations or realistic color distributions to improve
naturalness and black-box transferability.

More recently, diffusion models have attracted attention for their ability to
produce high-quality adversarial images. For example, \textit{DiffAttack}
projects an input image into the latent space through DDIM \cite{ddim}
inversion and manipulates latent representations or attention structures to
generate transferable adversarial examples \cite{diffattack}. Other studies
include \textit{AdvDiff}, which synthesizes unrestricted adversarial examples
through classifier-guided reverse diffusion \cite{ADVdiff}, and
\textit{NatADiff}, which guides the diffusion trajectory toward the intersection
between the true and adversarial classes to improve transferability
\cite{natadiff}.

These studies show that generative models provide an effective framework for
producing natural and transferable adversarial examples. However, many existing
methods rely on surrogate-model guidance and are therefore mainly studied in
white-box or transfer-based settings. In strict black-box settings, this can
misalign the attack direction with the victim model's decision boundary, making
them less effective than query-based attacks that optimize directly from
victim-model feedback.
\subsection{Prior-assisted query-based black-box attacks}

In query-based black-box attacks, the adversary optimizes perturbations solely through feedback from the victim model, without access to model parameters or gradients. While this setting is more realistic, directly exploring the attack space through queries alone can be inefficient, especially in high-dimensional input spaces. \textit{AdvFlow}, for example, employs a pre-trained normalizing flow and updates it through NES-based black-box queries \cite{AdvFlow}. While this is a highly realistic setting, it requires many victim-model queries. To improve query efficiency in black-box attacks, a growing body of work incorporates external priors into query-based optimization \cite{CG-attack, MCG-attack, difattack}. In contrast to purely query-driven attacks, these methods exploit information learned in advance---either from surrogate models or from generative models---and then adapt or apply it under black-box feedback.

One line of research uses transferable priors learned from surrogate models. For example, \textit{CGAttack} models a conditional adversarial distribution and improves black-box attack efficiency by partially transferring its learned parameters while updating the remaining components using target-model queries \cite{CG-attack}. \textit{MCGAttack} further extends this idea with a meta-learning framework, where a meta-generator trained across surrogate tasks can be rapidly adapted to a new target model with limited query feedback \cite{MCG-attack}. \textit{DifAttack} also leverages surrogate knowledge by training an autoencoder on clean--adversarial image pairs to disentangle adversarial and visual features, and then optimizing only the adversarial feature through black-box queries while keeping the visual feature fixed \cite{difattack}. Another line of research uses generative priors to improve query efficiency without explicitly relying on surrogate-model transfer.  
 However, many existing prior-assisted methods depend on surrogate-model transferability or additional training to construct attack-specific priors \cite{CG-attack, MCG-attack, difattack}. Moreover, most of them are developed under conventional perturbation-bounded ($\ell_p$-bounded) formulations, where the resulting adversarial examples may be less natural than unrestricted ones. This leaves the generation of NAEs under strict black-box access, without additional attack-specific training, relatively underexplored.

\section{Method}
\subsection{Problem definition}
\begin{wrapfigure}{r}{0.47\textwidth}
\vspace{-10pt}
\captionsetup{type=algorithm,labelfont=bf,labelsep=space,justification=raggedright,singlelinecheck=false}
\hrule height 0.8pt
\vspace{2pt}
\caption{Adversarial Scenario Attack}
\label{alg:asa}
\vspace{-6pt}
\hrule height 0.4pt

\footnotesize
\begin{algorithmic}[1]
\Require image $x$, label $y$, victim $F$, generative editor $G$, MLLM $\mathcal{M}$, budget $Q$
\Statex Each computation of $J(\cdot)$ consumes one query (q).

\For{$t=1,2,\ldots$ \textbf{while} $q<Q$}
    \State Generate $\{p_t^r\}_{r\in\mathcal{R}}$ using $\mathcal{M}(x,y,\mathcal{H})$
    \For{each $r\in\mathcal{R}$}
        \State $x_t^r\gets G(x,p_t^r)$
        \State $J(p_t^r)\gets\max_{c\ne y}F_c(x_t^r)-F_y(x_t^r)$
        \If{$J(p_t^r)>0$ and \textsc{Verify}$_{\mathcal{M}}(x,x_t^r,p_t^r)$}
            \State \Return $x_t^r$
        \EndIf
    \EndFor
    \State $r_t^+\gets\arg\max_{r\in\mathcal{R}}J(p_t^r)$, $p_t^+\gets p_t^{r_t^+}$
    \State $\mathcal{P}_t^-\gets\{p_t^r\mid r\in\mathcal{R}\setminus\{r_t^+\}\}$
    \State Update $\mathcal{H}$ with winner--loser feedback 
    \State $\mathcal{C}_t\gets\{p_t^r:r\in\mathcal{R}\}$, and add $p_{t-1}^g$ if $t>1$
    \State $g_t\gets\arg\max_{p\in\mathcal{C}_t}J(p)$
    \For{each remaining $u\in\mathcal{C}_t$ sorted by $J(u)$}
        \State $\hat{p}\gets g_t \oplus_{\mathrm{and}} u$
        \State $\hat{x}\gets G(x,\hat{p})$
        \State Compute $J(\hat{p})$
        \If{$J(\hat{p})>0$ and \textsc{Verify}$_{\mathcal{M}}(x,\hat{x},\hat{p})$}
            \State \Return $\hat{x}$
        \EndIf
        \If{$J(\hat{p})>J(g_t)$}
            \State $g_t\gets\hat{p}$
        \EndIf
    \EndFor
    \State $p_t^g\gets g_t$
\EndFor
\State \Return failure
\end{algorithmic}

\vspace{4pt}
\hrule height 0.8pt
\vspace{-15pt}
\end{wrapfigure}
We consider a score-based query black-box attack setting. Let 
\(F: \mathcal{X} \rightarrow \mathbb{R}^{C}\) denote a victim classifier, 
where \(C\) is the number of classes. Given an input image 
\(x \in \mathcal{X}\) with ground-truth label \(y\), the attacker has no access 
to the architecture, parameters, or gradients of \(F\). The attacker can only 
query \(F\) and observe its output confidence scores. 
Unlike conventional black-box attacks that directly search for perturbations
in the pixel space, we formulate the attack as a search over natural-language
editing scenarios. Let \(G\) be a generative image editor that takes an input
image \(x\) and a natural-language editing instruction \(p \in \mathcal{P}\),
and produces an edited image $x_p = G(x,p), $ where \(\mathcal{P}\) denotes the space of natural-language editing
instructions. Here, \(G(x,p)\) includes the internal text encoding and image
editing process of the generative model.

The goal of the attacker is to find an editing scenario \(p\) that induces a
realistic edit of the original image while causing the victim model to
misclassify it. Following the margin loss commonly used in adversarial
attacks~\cite{att_ref4}, we define the attack score as
$
J(p)=\mathcal{L}_{\mathrm{margin}}(x_p,y)
=\max_{c \neq y} F_c(x_p)-F_y(x_p),
$
where \(F_c(x_p)\) denotes the victim model's output confidence score for class
\(c\). A larger margin implies that \(x_p\) is closer to, or has already
crossed, the decision boundary; in particular, \(J(p)>0\) indicates a
successful attack. We focus on an extremely query-limited regime, allowing at
most 100 queries to the victim model for each input image. This budget is enforced before every
victim-model query. Under this
constraint, our goal is to efficiently identify adversarial editing scenarios
without relying on high-dimensional gradient estimation or task-specific
surrogate model training.

\subsection{Adversarial scenario attack}

Building on the problem formulation above, we propose Adversarial Scenario
Attack (ASA), a score-based black-box attack that searches over natural
language editing scenarios rather than pixel-level perturbations. ASA organizes
its search space by three editing strategies: Background Shift, Weather \&
Atmosphere, and Material \& Color Reskinning. We denote this strategy set as
\(\mathcal{R}=\{\mathrm{BG}, \mathrm{WA}, \mathrm{MCR}\}\). Each strategy
\(r \in \mathcal{R}\) defines a scenario space
\(\mathcal{S}_r \subset \mathcal{P}\), consisting of plausible natural language
editing instructions for that transformation type. The full ASA search space is
\(\mathcal{S}_{\mathrm{ASA}}=\bigcup_{r\in\mathcal{R}}\mathcal{S}_r\).
At each iteration, ASA evaluates candidates from all strategy-specific scenario
spaces under a shared query budget, enabling direct comparison among
heterogeneous natural transformations.

ASA proceeds as an iterative history-guided search. We maintain a search
history \(\mathcal{H}\), initialized as an empty set, that records previously
tested strategy-specific scenarios together with their attack scores and
verifier feedback when available. 
At iteration \(t\), a MLLM observes the source image \(x\), the ground-truth label \(y\), and
\(\mathcal{H}\). 
It then proposes one candidate scenario
\(p_t^r \in \mathcal{S}_r\) for each strategy \(r \in \mathcal{R}\), resulting
in a batch of strategy-specific candidates. 
The scenario generator is
constrained to produce benign-looking and visually plausible edits, while
avoiding explicit class injection, object replacement, or unnatural
corruptions. Thus, ASA searches for adversarial scenarios that resemble
ordinary user edit requests.

For each candidate scenario \(p_t^r\), ASA generates the edited image
\(x_t^r=G(x,p_t^r)\), queries the victim classifier \(F\), and assigns the
attack score \(J(p_t^r)=\mathcal{L}_{\mathrm{margin}}(x_t^r,y)\). We define
the winning strategy and scenario as
\(r_t^{+}=\arg\max_{r\in\mathcal{R}} J(p_t^r)\) and
\(p_t^{+}=p_t^{r_t^{+}}\), and the losing scenarios as
\(\mathcal{P}_t^{-}=\{p_t^r \mid r\in\mathcal{R}\setminus\{r_t^{+}\}\}\).
ASA updates the search history with the winning scenario, the losing
scenarios, their attack scores, and verifier feedback when available. In the
next proposal round, the MLLM is conditioned to exploit
\(p_t^{+}\) by generating more specific variants within \(r_t^{+}\), while
avoiding semantic traits shared by the losing scenarios in
\(\mathcal{P}_t^{-}\). If a candidate is misclassified by \(F\), it is passed
to the naturalness verifier; otherwise, the search continues until the
victim-query budget is exhausted.

To avoid degenerate successes caused by visibly corrupted or implausible
images, ASA uses a MLLM as a naturalness verifier. The
verifier receives a side-by-side comparison of the source and edited images,
the base description of the source image, and the requested scenario. It
evaluates whether the edited image is a plausible result of the requested
scenario, whether the main subject remains consistent with the base
description, and whether the edit is limited to intended factors such as
background, atmosphere, lighting, or surface appearance. If the verifier
rejects the edited image, ASA does not terminate; instead, it returns a short
feedback sentence describing the main issue, which is added to the search
history for the next proposal round. The final adversarial example is therefore
required to satisfy both classifier failure and visual plausibility. This verifier is designed to enforce label-preserving semantic validity: the edited image must preserve the main object while remaining a plausible result of the requested natural transformation.

\subsection{Greedy explorer}

To move beyond independently evaluating strategy-specific candidates, the
Greedy Explorer composes promising scenarios into a unified multi-scenario
prompt. This allows ASA to search for semantic vulnerabilities that may not be
exposed by any single strategy alone but emerge from the interaction of
multiple natural transformations. At iteration \(t\), ASA forms a candidate
pool from the current strategy-specific scenarios. From the second iteration
onward, the pool also includes the greedy scenario carried over from the
previous iteration:
\begin{equation}
\label{eq:greedy_pool}
\mathcal{C}_t =
\begin{cases}
\{p_t^r \mid r\in\mathcal{R}\}, & t=1,\\
\{p_t^r \mid r\in\mathcal{R}\}\cup\{p_{t-1}^g\}, & t>1.
\end{cases}
\end{equation}
The explorer selects the best-scoring scenario in \(\mathcal{C}_t\) as the
initial greedy state:
$
g_t^{(0)}=\arg\max_{p\in\mathcal{C}_t}J(p).
$
The remaining scenarios are obtained from
\(\mathcal{C}_t\setminus\{g_t^{(0)}\}\). When \(t>1\), if the previous greedy
scenario \(p_{t-1}^g\) remains in this set, ASA replaces it in the traversal list
with its strategy-keyed components \(\mathrm{Comp}(p_{t-1}^g)\), rather than
attaching the entire composed scenario at once. If \(g_t^{(0)}=p_{t-1}^g\), its
components are already represented in the initial greedy state and are not added
again. The remaining scenarios are then visited sequentially in descending order
of their individual attack scores. Let \(u_t^{(m)}\) denote the \(m\)-th
remaining element to be tested, and let \(g_t^{(m-1)}\) denote the current
greedy scenario before testing it. ASA forms a trial scenario by composing the
two instructions with a conjunction:
\begin{equation}
\label{eq:greedy_and}
\hat{p}_t^{(m)} = g_t^{(m-1)} \oplus_{\mathrm{and}} u_t^{(m)},
\end{equation}
where \(\oplus_{\mathrm{and}}\) denotes component-level natural-language
scenario composition. We represent each composed scenario as a set of
strategy-keyed components; when \(u_t^{(m)}\) has the same strategy key as a
component already present in \(g_t^{(m-1)}\), the existing component is replaced
with the newer one, and otherwise the new component is appended. The composed
scenario is then rendered as \(G(x,\hat{p}_t^{(m)})\) and evaluated by querying
the victim classifier.
The greedy state is updated only when the composed scenario improves the
current attack objective:
\begin{equation}
\label{eq:greedy_update}
g_t^{(m)}=
\begin{cases}
\hat{p}_t^{(m)}, & \text{if } J(\hat{p}_t^{(m)})>J(g_t^{(m-1)}),\\
g_t^{(m-1)}, & \text{otherwise}.
\end{cases}
\end{equation}
After all remaining scenarios are visited, the accumulated greedy scenario is
set to
$
p_t^g = g_t^{(M)},
$
where \(M\) is the number of visited scenarios.
This sequential update allows each score-improving composition to become the
reference for the next trial.

\section{Experiments}
\subsection{Experimental details}
\label{sec4.1}
\paragraph{Datasets.}
For query-based and transfer attack evaluation, we use an ImageNet-compatible
dataset consisting of 1,000 images~\cite{imagenet}. Each image is resized from
\(299 \times 299\) to \(224 \times 224\). For prompt-level transferability, we
use a separate subset of the ImageNet validation set~\cite{imagenet_val},
consisting of 430 classes with 50 images per class, resulting in 21,500 images.
This subset is used to evaluate the transferability of class-level scenarios
discovered in Sec.~\ref{query}.
\paragraph{Victim models.}
We evaluate ASA on a diverse set of ImageNet classifiers to assess its effectiveness under query-based black-box attacks.
The victim models include CNN-based architectures, ResNet-50 (RN-50) \cite{resnet} and ConvNeXt \cite{convnext}; Transformer-based architectures, Swin-Base \cite{swin} and DeiT-Base \cite{deit}; and state-space-model-based architectures, Vision Mamba-Small (ViM-S) \cite{vim} and MambaVision-Base \cite{mambavision}.
For robustness evaluation, we further consider adversarially trained models, including Adversarial Inception-v3 (Adv-Inc) and ensemble adversarially trained Inception-ResNet-v2 (Adv-Res) \cite{robust_models}.
In the transferability experiments, we additionally evaluate Inception-v3 (Inc-v3) \cite{inception}, Wide-ResNet-50 (WRN-50)~\cite{wrn} and Vision Transformer-Base (ViT-B) \cite{vit}.
\paragraph{Implementation details.}
We use Gemma 4 E4B \cite{gemma} as the MLLM for scenario generation and
naturalness evaluation, without fine-tuning or task-specific adaptation
(\texttt{thinking=false}, \texttt{max\_new\_tokens}=4096,
\texttt{do\_sample=false}). For text-guided image editing, we use FLUX.2 [klein]
9B-KV \cite{flux-2-klein} with \(1024 \times 1024\) inputs and seed 123. For
each strategy, the MLLM generates one scenario per iteration; search runs for
up to 50 iterations and stops once victim-model queries reach 100. All experiments were conducted on a Linux server running Ubuntu 22.04 with an NVIDIA B200 GPU and an AMD EPYC 9365 CPU. Detailed prompts are
provided in App.~\ref{app:detail_prompts}.

\subsection{Comparison}
\label{sec4.2}
\subsubsection{Query-based attacks}
\label{query}
\begin{table*}[h]
\setlength{\tabcolsep}{0.9mm}
\centering
\caption{Query-based attack results. Higher ASR (\%) and lower Avg. Q (successful samples only) indicate better performance. Best results in \textcolor{red}{\textbf{red bold}}.}
\vspace{-2mm}
\label{tab:query_attack}
\scalebox{.74}{%
\begin{tabular}{lcccccccccccccccc}
\toprule[1.5pt]
\multirow{3}{*}{Methods} & \multicolumn{4}{c}{CNNs} & \multicolumn{4}{c}{Transformers} & \multicolumn{4}{c}{SSMs} & \multicolumn{4}{c}{Robust Models} \\
\cmidrule(lr){2-5} \cmidrule(lr){6-9} \cmidrule(lr){10-13} \cmidrule(lr){14-17}
& \multicolumn{2}{c}{ResNet-50} & \multicolumn{2}{c}{ConvNeXt}
& \multicolumn{2}{c}{Swin} & \multicolumn{2}{c}{DeiT}
& \multicolumn{2}{c}{ViM} & \multicolumn{2}{c}{MambaVision}
& \multicolumn{2}{c}{Adv-Res} & \multicolumn{2}{c}{Adv-Inc} \\
\cmidrule(lr){2-3} \cmidrule(lr){4-5} \cmidrule(lr){6-7} \cmidrule(lr){8-9}
\cmidrule(lr){10-11} \cmidrule(lr){12-13} \cmidrule(lr){14-15} \cmidrule(lr){16-17}
& ASR & Avg Q & ASR & Avg Q & ASR & Avg Q & ASR & Avg Q & ASR & Avg Q & ASR & Avg Q & ASR & Avg Q & ASR & Avg Q \\
\midrule
AdvFlow
& 18.56 & 18.28 & 5.83 & 32.30 & 4.08 & 24.00 & 5.93 & 27.54
& 8.14 & 29.93 & 3.42 & 29.33 & 17.47 & 23.29 & 22.55 & 20.45 \\
DIFAttack
& 46.67 & 36.99 & 17.18 & 46.43 & 23.93 & 45.81 & 25.42 & 48.42
& 26.45 & 43.44 & 17.22 & 51.99 & 27.52 & 36.83 & 36.20 & 30.47 \\
CGAttack
& 43.99 & 18.11 & 22.19 & 16.08 & 19.54 & 26.06 & 16.31 & 35.32
& 17.67 & 19.13 & 13.38 & 25.90 & 31.10 & 30.07 & 51.82 & 16.92 \\
MCGAttack
& 82.62 & \best{4.16} & 49.39 & 29.96 & 23.82 & 30.78 & 26.59 & 35.20
& 32.12 & 29.48 & 20.85 & 34.67 & 44.30 & 24.24 & 52.78 & 22.10 \\
ASA(ours)
& \best{94.10} & 9.10 & \best{87.22} & \best{13.76} & \best{89.34} & \best{12.76} & \best{88.03} & \best{11.73}
& \best{90.90} & \best{11.23} & \best{77.90} & \best{14.48} & \best{95.87} & \best{7.75} & \best{97.69} & \best{6.11} \\
\bottomrule[1.5pt]
\vspace{-1cm}
\end{tabular}
}
\end{table*}
Here, we compare ASA with state-of-the-art prior-assisted query-based
generative attacks, including AdvFlow~\cite{AdvFlow},
DIFAttack~\cite{difattack}, CGAttack~\cite{CG-attack}, and
MCGAttack~\cite{MCG-attack}. We report the Attack Success Rate (ASR) and the
average number of victim-model queries (Avg. Q) under the same query budget.
To avoid ASR inflation from prompt leakage, we count predictions of either the
original class or any explicitly leaked ImageNet category as failures; leakage
occurs in only 0.9--6.5\% of ASA-generated scenarios, with details in
App.~\ref{app:limitation}. Tab.~\ref{tab:query_attack} shows that ASA achieves the highest ASR on every
evaluated victim model and the lowest Avg. Q on seven of the eight models.
This advantage is especially pronounced on modern and heterogeneous model
families, where prior query-based attacks often trade off attack
success against query efficiency. These results show that scenario-level search
offers a practical, architecture-agnostic alternative for black-box NAE
generation.

\subsubsection{Image-level transferability}
\label{transfer}
\begin{table*}[ht]
\centering
\small
\setlength{\tabcolsep}{3.8pt}
\caption{ASR (\%), Perc. Sim., and FID across three source models. $^{*}$ marks white-box results, and gray-shaded rows indicate query-based attacks. \protect\best{Bold}, \protect\second{underlined}, and \protect\third{red} values mark the best, second-best, and third-best ASR within each source model.}
\label{tab:transferability}
\vspace{-2mm}
\scalebox{.68}{%
\begin{tabular}{ll*{12}{c}|c|c}
\toprule[1.5pt]
\multirow{2}{*}{\makecell{Source\\ Model}} & \multirow{2}{*}{Attack} & \multicolumn{4}{c}{CNNs} & \multicolumn{3}{c}{Transformers} & \multicolumn{2}{c}{SSMs} & \multicolumn{2}{c}{Robust Models} & \multirow{2}{*}{\makecell{Avg.\\ ASR}} & \multirow{2}{*}{\makecell{Perc.\\Sim.$\uparrow$}} & \multirow{2}{*}{\makecell{FID$\downarrow$}} \\
\cmidrule(lr){3-6} \cmidrule(lr){7-9} \cmidrule(lr){10-11} \cmidrule(lr){12-13}
 & & RN-50 & WRN-50 & Inc-v3 & ConvNeXt & ViT & Swin & DeiT & ViM & MambaVision & Adv-Inc & Adv-Res & & & \\
\midrule
\multirow{9}{*}{Swin} & CF & 45.5 & 41.8 & 30.2 & 13.5 & 27.3 & 60.6$^{*}$ & 18.2 & 22.6 & 10.2 & 37.8 & 29.5 & 30.7 & 0.95 & 48.55 \\
 & NCF & 45.7 & 40.7 & 35.1 & 28.1 & 31.1 & \third{83.6$^{*}$} & 27.1 & 27.8 & 21.1 & 30.7 & 28.0 & 36.3 & 0.96 & 41.47 \\
 & ACA & \best{60.3} & \best{59.0} & \best{62.3} & \second{57.0} & \best{58.6} & 80.8$^{*}$ & \second{59.2} & \best{60.9} & \second{58.0} & \best{65.9} & \best{62.2} & \best{62.2} & 0.53 & 66.18 \\
 & DIFFAttack & \third{54.1} & \second{56.5} & \third{46.8} & \best{67.2} & \second{58.1} & \best{89.8$^{*}$} & \best{63.9} & \second{56.8} & \best{65.5} & \third{44.6} & \third{39.7} & \second{58.5} & 0.75 & 52.66 \\
 & \blackcell{AdvFlow} & \blackcell{6.1} & \blackcell{7.0} & \blackcell{10.5} & \blackcell{1.2} & \blackcell{2.5} & \blackcell{5.1} & \blackcell{1.8} & \blackcell{1.9} & \blackcell{0.7} & \blackcell{8.7} & \blackcell{7.9} & \blackcell{4.9} & \blackcell{0.98} & \blackcell{22.58} \\
 & \blackcell{DIFAttack} & \blackcell{12.8} & \blackcell{10.2} & \blackcell{14.1} & \blackcell{4.5} & \blackcell{4.6} & \blackcell{23.9} & \blackcell{4.5} & \blackcell{5.3} & \blackcell{2.2} & \blackcell{16.7} & \blackcell{9.6} & \blackcell{9.9} & \blackcell{0.96} & \blackcell{37.29} \\
 & \blackcell{CGAttack} & \blackcell{17.6} & \blackcell{16.4} & \blackcell{16.5} & \blackcell{10.3} & \blackcell{4.8} & \blackcell{19.5} & \blackcell{4.3} & \blackcell{6.5} & \blackcell{5.8} & \blackcell{35.8} & \blackcell{13.5} & \blackcell{13.7} & \blackcell{0.96} & \blackcell{71.54} \\
 & \blackcell{MCGAttack} & \blackcell{37.7} & \blackcell{39.1} & \blackcell{42.2} & \blackcell{12.5} & \blackcell{9.8} & \blackcell{23.8} & \blackcell{6.5} & \blackcell{9.3} & \blackcell{5.0} & \blackcell{27.9} & \blackcell{16.9} & \blackcell{21.0} & \blackcell{0.94} & \blackcell{75.58} \\
 & \blackcell{ASA (ours)} & \blackcell{\second{56.1}} & \blackcell{\third{56.0}} & \blackcell{\second{57.3}} & \blackcell{\third{47.1}} & \blackcell{\third{51.9}} & \blackcell{\second{89.3}} & \blackcell{\third{50.7}} & \blackcell{\third{53.1}} & \blackcell{\third{36.7}} & \blackcell{\second{62.3}} & \blackcell{\second{54.6}} & \blackcell{\third{55.9}} & \blackcell{0.78} & \blackcell{60.64} \\
\midrule
\multirow{9}{*}{ConvNeXt} & CF & 42.5 & 42.0 & 26.5 & 49.8$^{*}$ & 25.6 & 17.4 & 18.5 & 21.2 & 12.3 & 38.6 & 29.4 & 29.4 & 0.95 & 48.77 \\
 & NCF & 51.3 & 49.0 & 44.4 & 66.2$^{*}$ & 38.3 & 34.6 & 30.6 & 32.5 & 23.9 & 36.8 & 32.4 & 40.0 & 0.95 & 46.89 \\
 & ACA & \second{63.8} & \second{64.9} & \second{65.3} & \third{80.4$^{*}$} & \second{60.8} & \second{66.2} & \second{62.5} & \second{60.9} & \second{60.3} & \best{66.1} & \best{63.6} & \second{65.0} & 0.51 & 67.47 \\
 & DIFFAttack & \best{78.9} & \best{81.4} & \best{67.1} & \best{97.6$^{*}$} & \best{72.5} & \best{88.2} & \best{78.4} & \best{73.1} & \best{83.8} & \third{58.9} & \second{59.4} & \best{76.3} & 0.63 & 71.82 \\
 & \blackcell{AdvFlow} & \blackcell{7.5} & \blackcell{8.0} & \blackcell{12.4} & \blackcell{6.3} & \blackcell{2.3} & \blackcell{1.3} & \blackcell{2.2} & \blackcell{2.4} & \blackcell{0.7} & \blackcell{10.6} & \blackcell{7.7} & \blackcell{5.6} & \blackcell{0.98} & \blackcell{23.40} \\
 & \blackcell{DIFFAttack} & \blackcell{14.3} & \blackcell{11.3} & \blackcell{15.7} & \blackcell{17.2} & \blackcell{4.6} & \blackcell{5.0} & \blackcell{4.0} & \blackcell{4.9} & \blackcell{3.0} & \blackcell{18.0} & \blackcell{10.4} & \blackcell{9.9} & \blackcell{0.96} & \blackcell{36.89} \\
 & \blackcell{CGAttack} & \blackcell{21.2} & \blackcell{18.7} & \blackcell{16.7} & \blackcell{22.2} & \blackcell{4.3} & \blackcell{8.3} & \blackcell{4.7} & \blackcell{8.4} & \blackcell{6.2} & \blackcell{34.0} & \blackcell{14.3} & \blackcell{14.5} & \blackcell{0.95} & \blackcell{83.89} \\
 & \blackcell{MCGAttack} & \blackcell{41.4} & \blackcell{42.4} & \blackcell{43.0} & \blackcell{49.4} & \blackcell{9.5} & \blackcell{6.0} & \blackcell{5.8} & \blackcell{10.0} & \blackcell{5.0} & \blackcell{27.9} & \blackcell{18.0} & \blackcell{23.5} & \blackcell{0.93} & \blackcell{76.91} \\
 & \blackcell{ASA (ours)} & \blackcell{\third{59.3}} & \blackcell{\third{56.9}} & \blackcell{\third{57.3}} & \blackcell{\second{87.2}} & \blackcell{\third{53.1}} & \blackcell{\third{50.7}} & \blackcell{\third{48.1}} & \blackcell{\third{52.4}} & \blackcell{\third{40.5}} & \blackcell{\second{63.3}} & \blackcell{\third{56.8}} & \blackcell{\third{56.9}} & \blackcell{0.78} & \blackcell{62.04} \\
\midrule
\multirow{9}{*}{ViM} & CF & 39.7 & 35.3 & 24.1 & 9.2 & 18.6 & 10.2 & 12.9 & 69.4$^{*}$ & 6.8 & 35.2 & 26.6 & 26.2 & 0.96 & 43.03 \\
 & NCF & 46.9 & 42.2 & 37.9 & 25.0 & 36.4 & 25.7 & 30.5 & 77.7$^{*}$ & 19.7 & 32.3 & 29.4 & 36.7 & 0.95 & 43.93 \\
 & ACA & \best{59.7} & \best{59.0} & \best{62.5} & \best{53.8} & \second{60.4} & \second{57.1} & \second{60.8} & \third{84.5$^{*}$} & \second{52.2} & \best{66.9} & \best{61.2} & \best{61.6} & 0.52 & 65.16 \\
 & DIFFAttack & \third{51.0} & \third{51.7} & \third{51.7} & \second{49.7} & \best{62.7} & \best{57.1} & \best{63.4} & \second{88.9$^{*}$} & \best{52.7} & \third{44.5} & \third{40.4} & \second{55.8} & 0.76 & 51.89 \\
 & \blackcell{AdvFlow} & \blackcell{7.0} & \blackcell{6.7} & \blackcell{9.9} & \blackcell{1.3} & \blackcell{3.2} & \blackcell{0.8} & \blackcell{2.0} & \blackcell{9.6} & \blackcell{0.9} & \blackcell{9.4} & \blackcell{7.4} & \blackcell{5.3} & \blackcell{0.98} & \blackcell{22.05} \\
 & \blackcell{DIFAttack} & \blackcell{13.5} & \blackcell{9.7} & \blackcell{14.1} & \blackcell{2.3} & \blackcell{4.2} & \blackcell{4.0} & \blackcell{4.6} & \blackcell{26.5} & \blackcell{2.0} & \blackcell{18.2} & \blackcell{7.9} & \blackcell{9.7} & \blackcell{0.96} & \blackcell{37.33} \\
 & \blackcell{CGAttack} & \blackcell{21.4} & \blackcell{17.9} & \blackcell{16.8} & \blackcell{9.5} & \blackcell{3.2} & \blackcell{6.7} & \blackcell{3.8} & \blackcell{17.8} & \blackcell{5.0} & \blackcell{32.9} & \blackcell{10.0} & \blackcell{13.2} & \blackcell{0.96} & \blackcell{95.35} \\
 & \blackcell{MCGAttack} & \blackcell{38.2} & \blackcell{38.6} & \blackcell{41.9} & \blackcell{13.4} & \blackcell{10.2} & \blackcell{5.5} & \blackcell{5.4} & \blackcell{32.1} & \blackcell{3.8} & \blackcell{27.9} & \blackcell{16.3} & \blackcell{21.2} & \blackcell{0.94} & \blackcell{77.93} \\
 & \blackcell{ASA (ours)} & \blackcell{\second{53.7}} & \blackcell{\second{52.0}} & \blackcell{\second{57.1}} & \blackcell{\third{39.1}} & \blackcell{\third{48.7}} & \blackcell{\third{43.8}} & \blackcell{\third{46.9}} & \blackcell{\best{90.9}} & \blackcell{\third{30.9}} & \blackcell{\second{61.6}} & \blackcell{\second{55.4}} & \blackcell{\third{52.7}} & \blackcell{0.80} & \blackcell{59.94} \\
\bottomrule[1.5pt]
\vspace{-0.8cm}
\end{tabular}%
}
\end{table*}

Tab.~\ref{tab:transferability} compares ASA with prior-assisted query-based
generative attacks and representative transfer-based attacks, including
CF~\cite{CF}, NCF~\cite{NCF}, ACA~\cite{ACA}, and DIFFAttack~\cite{diffattack}.
The table evaluates whether adversarial images generated from a source model
remain effective across heterogeneous victim architectures. To evaluate visual fidelity, we report DINOv2-based perceptual similarity~\cite{dinov2,stable-flow}
and Fr\'echet Inception Distance (FID)~\cite{fid}. Higher perceptual similarity
indicates better semantic preservation, while lower FID indicates closer
distributional similarity to the original images.
\textbf{Among query-based attacks, ASA consistently shows the strongest image
transferability across all source models.} Compared with prior-assisted
query-based baselines, ASA produces adversarial images that transfer much more
reliably to CNNs, Transformers, SSMs, and robust models. This indicates that
scenario-level adversarial generation is not merely effective for the queried
model, but also captures transformations that generalize across architectures.

\textbf{ASA is also competitive with transfer-based attacks that explicitly exploit
source-model supervision.} Although strong transfer attacks such as ACA and
DIFFAttack achieve high ASR in some settings, ASA remains close to them while
using only query-based feedback. While ACA achieves the highest ASR on robust victim models, ASA consistently
remains competitive and ranks within the top three across source models. More importantly, among these highly transferable attacks, ASA achieves the best
DINOv2-based perceptual similarity and maintains competitive FID scores,
especially improving over ACA in distributional fidelity, indicating a better
balance between robust transferability and visual fidelity.
Overall, the image transferability results demonstrate that ASA combines the
practicality of query-based attacks with transferability comparable to strong
transfer-based methods, making it effective across diverse model families.

\subsection{Prompt-level transferability}
\label{sec:prompt_transfer}
\begin{table*}[ht]
\centering
\small
\setlength{\tabcolsep}{3.8pt}
\caption{Prompt transferability of ASA and its ablations measured by ASR (\%)
across victim models. \(w/o~S,G\) removes structured scenario guidance
and greedy exploration; \(w/o~G\) removes only greedy exploration.
\protect\best{Red bold} denotes the best result, and $^{*}$ marks source-model
evaluation.}
\label{tab:prompt_transfer_asa}
\vspace{-2mm}
\scalebox{.74}{%
\begin{tabular}{ll*{12}{c}}
\toprule[1.5pt]
\multirow{3}{*}{\makecell{Source\\ Model}} & \multirow{3}{*}{Method} & \multicolumn{11}{c}{Victim Models} & \multirow{3}{*}{\makecell{Avg.\\ ASR}} \\
\cmidrule(lr){3-13}
 & & \multicolumn{4}{c}{CNNs} & \multicolumn{3}{c}{Transformers} & \multicolumn{2}{c}{SSMs} & \multicolumn{2}{c}{Robust Models} & \\
\cmidrule(lr){3-6} \cmidrule(lr){7-9} \cmidrule(lr){10-11} \cmidrule(lr){12-13}
 & & RN-50 & WRN-50 & Inc-v3 & ConvNeXt & ViT & Swin & DeiT & ViM & MambaVision & Adv-Inc & Adv-Res & \\
\midrule
\multirow{3}{*}{Swin} & ASA$_{\mathrm{w/o}\,S,G}$ & 33.42 & 31.89 & 38.36 & 25.13 & 27.97 & 27.47$^{*}$ & 27.40 & 28.68 & 25.06 & 41.30 & 35.66 & 31.12 \\
 & ASA$_{\mathrm{w/o}\,G}$ & 40.26 & 37.44 & 44.15 & 26.74 & 32.40 & 29.53$^{*}$ & 30.13 & 32.43 & 24.67 & 48.77 & 41.42 & 35.27 \\
 & ASA & \best{49.45} & \best{46.55} & \best{53.65} & \best{33.82} & \best{40.32} & \best{36.68}$^{*}$ & \best{37.49} & \best{40.60} & \best{31.55} & \best{58.00} & \best{49.70} & \best{43.44} \\
\midrule
\multirow{3}{*}{ConvNeXt} & ASA$_{\mathrm{w/o}\,S,G}$ & 33.79 & 32.52 & 38.55 & 25.74$^{*}$ & 28.54 & 27.24 & 27.60 & 29.13 & 25.17 & 40.63 & 35.80 & 31.34 \\
 & ASA$_{\mathrm{w/o}\,G}$ & 43.22 & 39.91 & 46.46 & 30.06$^{*}$ & 34.47 & 30.68 & 30.80 & 34.40 & 25.69 & 51.40 & 43.87 & 37.36 \\
 & ASA & \best{50.06} & \best{47.01} & \best{53.93} & \best{34.64}$^{*}$ & \best{41.15} & \best{36.94} & \best{38.00} & \best{40.81} & \best{32.13} & \best{57.77} & \best{49.76} & \best{43.84} \\
\midrule
\multirow{3}{*}{ViM} & ASA$_{\mathrm{w/o}\,S,G}$ & 28.80 & 27.46 & 33.66 & 20.57 & 24.13 & 22.47 & 23.01 & 24.40$^{*}$ & 20.37 & 37.04 & 31.77 & 26.70 \\
 & ASA$_{\mathrm{w/o}\,G}$ & 40.10 & 37.68 & 44.66 & 26.34 & 32.69 & 28.50 & 30.04 & 33.30$^{*}$ & 25.37 & 49.28 & 42.05 & 35.46 \\
 & ASA & \best{47.32} & \best{44.40} & \best{52.00} & \best{31.49} & \best{39.24} & \best{34.86} & \best{36.46} & \best{39.41}$^{*}$ & \best{30.68} & \best{55.60} & \best{48.25} & \best{41.79} \\
\bottomrule[1.5pt]
\vspace{-0.8cm}
\end{tabular}%
}
\end{table*}
We introduce prompt-level transferability to evaluate whether a class-level
adversarial editing scenario discovered on a source model remains effective for
unseen images of the same class and for victim models with different
architectures. This protocol differs from conventional image-level
transferability: it evaluates reusable natural-language prompts rather than
fixed adversarial images or perturbations. \textbf{This distinction makes direct
comparison with prior attacks unsuitable}, since prior query-based or
transfer-based attacks do not produce class-level reusable editing scenarios.
We therefore use ASA ablations as controlled baselines:
ASA$_{\mathrm{w/o}\,S,G}$ removes scenario guidance, winner-loser feedback, and
greedy composition, while ASA$_{\mathrm{w/o}\,G}$ removes only greedy
composition.
Given a source model and an ASA variant, we reuse the adversarial prompts
generated in Sec.~\ref{query}. Each prompt is kept unchanged and applied to a
separate ImageNet validation subset containing the same 430 ImageNet-compatible
classes, with 50 images per class, resulting in 21,500 images. We do not
re-optimize prompts for these validation images.

We compute ASR in a coverage-normalized manner. For each source-victim pair, we
count post-attack samples misclassified by the victim model for labels where the
method produces an adversarial prompt. Labels for which the source attack fails
to produce a prompt are counted as zero transferred successes, and the total is
divided by all 21,500 validation images. Tab.~\ref{tab:prompt_transfer_asa} shows that full ASA consistently improves
prompt transferability over its ablations across diverse victim models. This indicates that ASA does more than find image-specific
editing failures: it identifies class-level natural transformation
vulnerabilities that can be reused on unseen images and remain effective across
architectures. The gains over both ablations further show that these reusable
semantic failure modes arise from structured scenario search rather than generic
prompt perturbations alone.

\subsection{Image quality}
\label{sec4.4}
\begin{wraptable}{r}{0.44\textwidth}
\centering
\setlength{\tabcolsep}{1pt}
\vspace{-4mm}
\caption{Image quality assessment.}
\vspace{-2mm}
\label{tab:image_quality_no_fid}
\scalebox{.7}{%
\begin{tabular}{lccccc}
\toprule[1.5pt]
Attack &
\makecell{NIMA\\-AVA$\uparrow$} &
HyperIQA$\uparrow$ &
\makecell{MUSIQ\\-AVA$\uparrow$} &
\makecell{MUSIQ\\-KoniQ$\uparrow$} &
TReS$\uparrow$ \\
\midrule
CF         & \third{5.26}  & 0.561          & 4.00          & \third{51.80} & \third{71.40} \\
NCF        & 4.90          & 0.516          & 3.79          & 49.00          & 66.26 \\
DIFFAttack & 5.19          & \best{0.616}   & \second{4.16} & \best{55.61}  & \best{81.51} \\
ACA        & \best{5.38}   & \second{0.605} & \best{4.34}   & \second{55.51} & \second{79.11} \\
\rowcolor{blackrowgray}
AdvFlow    & 4.74          & 0.477          & 3.76          & 47.88          & 64.77 \\

\rowcolor{blackrowgray}
DIFAttack  & 4.43          & 0.448          & 3.68          & 39.71          & 49.38 \\
\rowcolor{blackrowgray}
CGAttack   & 4.52          & 0.430          & 3.74          & 42.06          & 55.42 \\
\rowcolor{blackrowgray}
MCGAttack & 4.96          & 0.541          & 3.66          & 46.16          & 59.98 \\
\rowcolor{blackrowgray}
ASA        & \second{5.30} & \third{0.565}  & \third{4.08}  & 50.26          & 71.06 \\

\bottomrule[1.5pt]
\end{tabular}%
}
\vspace{-0.35cm}
\end{wraptable}
We assess naturalness from complementary perspectives. The MLLM-based verifier
checks whether the main object is preserved and whether the edit is semantically
plausible, while the DINOv2-based perceptual similarity and FID reported in
Sec.~\ref{sec4.2} quantify semantic preservation and distributional fidelity.
Here, we further evaluate perceptual image quality using five no-reference image
quality assessment (NR-IQA) metrics: NIMA-AVA~\cite{NIMA},
HyperIQA~\cite{hyperiqa}, MUSIQ-AVA, MUSIQ-KoniQ~\cite{musiq}, and
TReS~\cite{tres}.
Tab.~\ref{tab:image_quality_no_fid} shows that ASA maintains competitive visual
quality despite operating as a query-based attack. ASA ranks in the top three on
NIMA-AVA, HyperIQA, and MUSIQ-AVA, and remains competitive on TReS.
Qualitatively, Fig.~\ref{fig:prompt_transferability} shows that ASA primarily
induces natural scene-level changes, such as lighting, weather, background, and
material variations, whereas several baselines introduce visible noise, color
artifacts, or local distortions. These results suggest that ASA improves attack
effectiveness without substantially degrading visual naturalness. Since ASA searches only for adversarial editing scenarios and does not modify
the image generation process, its absolute image quality is bounded by the
underlying editor \(G\) and may further benefit from advances in text-guided
image editing models.

\subsection{Ablation study}
\label{sec4.5}
\begin{wrapfigure}{r}{0.56\textwidth}
    \vspace{-0.35cm}
    \centering
    \includegraphics[width=0.56\textwidth]{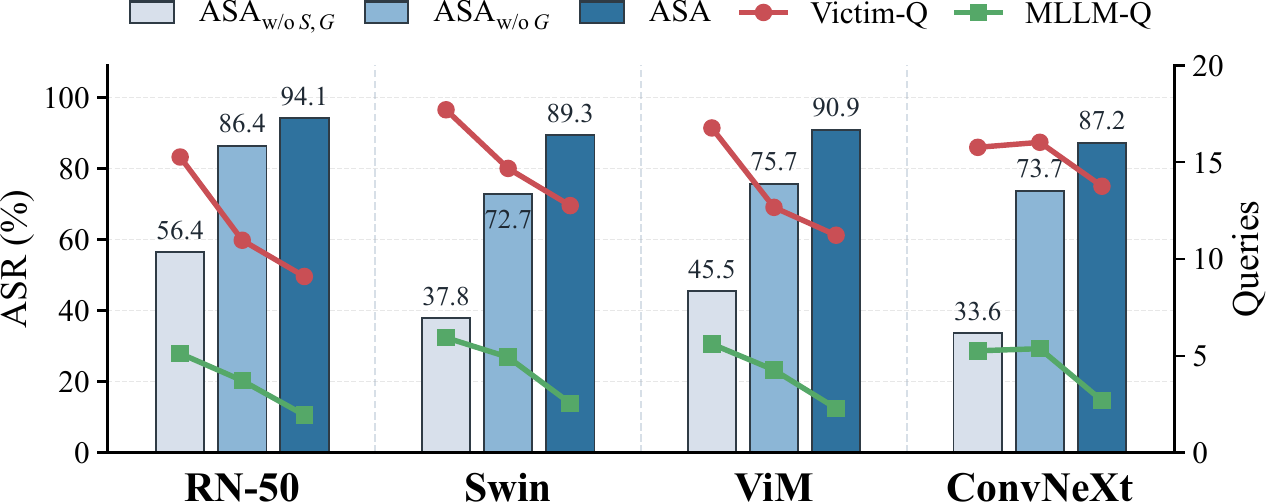}
    \vspace{-0.25cm}
    \caption{Ablation study of ASA components. Bars show ASR, while lines show victim-model and MLLM query counts across source models.}
    \label{fig:asa_ablation}
    \vspace{-0.3cm}
\end{wrapfigure}
Fig~\ref{fig:asa_ablation} analyzes the contribution of the two main components of ASA: scenario generation and greedy selection. Without both components, the attack shows a clear degradation in ASR, confirming that naive prompt construction is insufficient for reliable adversarial scenario generation.
Introducing scenario generation, denoted as ASA$_{\mathrm{w/o}\,G}$, substantially improves attack effectiveness, showing that semantically meaningful scenarios play a central role in producing transferable adversarial images. Adding greedy selection further strengthens ASA by selecting more effective scenarios while avoiding unnecessary queries. As a result, full ASA consistently achieves the best ASR across source models and also reduces both victim-model and MLLM query counts compared with ASA$_{\mathrm{w/o}\,G}$.
Overall, the ablation study demonstrates that the two components are complementary: scenario generation provides strong candidate attacks, while greedy selection improves both effectiveness and query efficiency. This confirms that the full ASA design is necessary for achieving strong and practical attack performance.

\subsection{Time analysis}

\label{sec4.6}
\begin{wraptable}{r}{0.4\textwidth}
\vspace{-0.4cm}
\centering
\small
\setlength{\tabcolsep}{0.9mm}
\caption{Runtime comparison.}
\vspace{-0.2cm}
\label{tab:runtime_comparison}
\begin{tabular}{lccc}
\toprule[1.5pt]
Attack & Prior (s) & Gen. (s) & Rel. \\
\midrule
AdvFlow   & 651.6   & 132.05  & 5.08$\times$ \\
DIFAttack & 4939.2  & 197.5   & 7.60$\times$ \\
CGAttack  & 15645.6 & 729.5   & 28.08$\times$ \\
MCGAttack       & 69343.2 & 1051.8  & 40.48$\times$ \\
ASA       & \best{0} & \best{25.98} & \best{1.00$\times$} \\
\bottomrule[1.5pt]
\vspace{-0.8cm}
\end{tabular}
\end{wraptable}
Tab.~\ref{tab:runtime_comparison} compares pre-generation training cost,
per-image generation time, and relative generation time. Here, Prior denotes
the additional attack-specific training required before generation, such as
learning an adversarial data distribution or training a surrogate model. Gen.
denotes the average generation time per image, and Rel. denotes the
generation-time ratio normalized by ASA. ASA achieves the most efficient runtime
profile: it requires no prior training and attains the shortest per-image
generation time among all compared query-based methods.

\section{Conclusion}
\label{5}
We proposed \textbf{Adversarial Scenario Attack (ASA)}, a query-based black-box framework for generating natural adversarial examples through natural-language scenario search. Instead of relying on surrogate gradients, attack-specific learned priors, or additional generative-model training, ASA exploits modern text-guided editing to explore realistic transformations such as background, weather, and material/color changes. By combining MLLM-guided scenario generation with a Greedy Explorer that composes attack-improving scenarios, ASA efficiently discovers natural edits that fool diverse vision models under tight query budgets. Our experiments show that ASA achieves strong attack success, low query cost, competitive image-level transferability, and prompt-level transferability across images and architectures. These results suggest that vision models exhibit reusable vulnerabilities to specific natural transformation patterns, highlighting scenario-level search as a practical direction for black-box robustness evaluation beyond conventional perturbation-centered attacks. Future work will investigate why such vulnerabilities arise and how the discovered scenarios can be incorporated into adversarial training to improve robustness against realistic semantic variations.

\medskip
\bibliographystyle{unsrt}
\bibliography{reference}

@inproceedings{att_ref1,
  author    = {Christian Szegedy and Wojciech Zaremba and Ilya Sutskever and Joan Bruna and Dumitru Erhan and Ian J. Goodfellow and Rob Fergus},
  title     = {{Intriguing properties of neural networks}},
  booktitle = {International Conference on Learning Representations},
  year      = {2014},
  doi       = {10.48550/arxiv.1312.6199}
}

@inproceedings{att_ref2,
  author    = {Battista Biggio and Igino Corona and Davide Maiorca and Blaine Nelson and Nedim Srndic and Pavel Laskov and Giorgio Giacinto and Fabio Roli},
  title     = {{Evasion Attacks against Machine Learning at Test Time}},
  booktitle = {European Conference on Machine Learning and Principles and Practice of Knowledge Discovery in Databases},
  year      = {2013},
  doi       = {10.1007/978-3-642-40994-3_25}
}

@inproceedings{att_ref3,
  author    = {Ian J. Goodfellow and Jonathon Shlens and Christian Szegedy},
  title     = {{Explaining and Harnessing Adversarial Examples}},
  booktitle = {International Conference on Learning Representations},
  year      = {2015},
  doi       = {10.48550/arxiv.1412.6572}
}

@inproceedings{att_ref4,
  author    = {Nicholas Carlini and David A. Wagner},
  title     = {{Towards Evaluating the Robustness of Neural Networks}},
  booktitle = {IEEE Symposium on Security and Privacy},
  year      = {2017},
  doi       = {10.1109/SP.2017.49}
}

@inproceedings{pgd,
  author    = {Aleksander Madry and Aleksandar Makelov and Ludwig Schmidt and Dimitris Tsipras and Adrian Vladu},
  title     = {{Towards Deep Learning Models Resistant to Adversarial Attacks}},
  booktitle = {International Conference on Learning Representations},
  year      = {2018},
  doi       = {10.48550/arxiv.1706.06083}
}

@inproceedings{ffsgm,
  author    = {Eric Wong and Leslie Rice and J. Zico Kolter},
  title     = {{Fast is better than free: Revisiting adversarial training}},
  booktitle = {International Conference on Learning Representations},
  year      = {2020},
  doi       = {10.48550/arxiv.2001.03994}
}

@inproceedings{rfpar,
  author    = {Dongsu Song and Daehwa Ko and JayHoon Jung},
  title     = {{Amnesia as a Catalyst for Enhancing Black Box Pixel Attacks in Image Classification and Object Detection}},
  booktitle = {Advances in Neural Information Processing Systems},
  year      = {2024},
  doi       = {10.52202/079017-2766}
}

@inproceedings{NCF,
  author       = {Shengming Yuan and
                  Qilong Zhang and
                  Lianli Gao and
                  Yaya Cheng and
                  Jingkuan Song},
  title        = {Natural Color Fool: Towards Boosting Black-box Unrestricted Attacks},
  booktitle    = {Advances in Neural Information Processing Systems},
  year         = {2022}
}

@inproceedings{ACA,
  author       = {Zhaoyu Chen and
                  Bo Li and
                  Shuang Wu and
                  Kaixun Jiang and
                  Shouhong Ding and
                  Wenqiang Zhang},
  title        = {Content-based Unrestricted Adversarial Attack},
  booktitle    = {Advances in Neural Information Processing Systems},
  year         = {2023}
}

@inproceedings{advdiffuser,
  author       = {Xinquan Chen and
                  Xitong Gao and
                  Juanjuan Zhao and
                  Kejiang Ye and
                  Cheng{-}Zhong Xu},
  title        = {AdvDiffuser: Natural Adversarial Example Synthesis with Diffusion
                  Models},
  booktitle    = {International Conference on Computer Vision},
  year         = {2023},
  doi          = {10.1109/ICCV51070.2023.00421}
}

@inproceedings{natadiff,
  author       = {Max Collins and
                  Jordan Vice and
                  Tim French and
                  Ajmal Mian},
  title        = {NatADiff: Adversarial Boundary Guidance for Natural Adversarial Diffusion},
  booktitle      = {International Conference on Learning Representations},
  year         = {2026},
  doi          = {10.48550/ARXIV.2505.20934},
}

@inproceedings{ACGAN,
  author       = {Yang Song and
                  Rui Shu and
                  Nate Kushman and
                  Stefano Ermon},
  title        = {Constructing Unrestricted Adversarial Examples with Generative Models},
  booktitle    = {Advances in Neural Information Processing Systems},
  year         = {2018}
}

@inproceedings{NAE1,
  author       = {Zhengli Zhao and
                  Dheeru Dua and
                  Sameer Singh},
  title        = {Generating Natural Adversarial Examples},
  booktitle    = {International Conference on Learning Representations},
  year         = {2018},
}

@inproceedings{ADVdiff,
  author       = {Xuelong Dai and
                  Kaisheng Liang and
                  Bin Xiao},
  title        = {AdvDiff: Generating Unrestricted Adversarial Examples Using Diffusion
                  Models},
  booktitle    = {European Conference on Computer Vision},
  year         = {2024},
  doi          = {10.1007/978-3-031-72952-2\_6},
}

@article{diffattack,
  author       = {Jianqi Chen and
                  Hao Chen and
                  Keyan Chen and
                  Yilan Zhang and
                  Zhengxia Zou and
                  Zhenwei Shi},
  title        = {Diffusion Models for Imperceptible and Transferable Adversarial Attack},
  journal      = {{IEEE} Transactions on Pattern Analysis and Machine Intelligence},
  volume       = {47},
  number       = {2},
  pages        = {961--977},
  year         = {2025},
  doi          = {10.1109/TPAMI.2024.3480519},
}

@inproceedings{NAE2,
  author       = {Dan Hendrycks and
                  Kevin Zhao and
                  Steven Basart and
                  Jacob Steinhardt and
                  Dawn Song},
  title        = {Natural Adversarial Examples},
  booktitle    = {Conference on Computer Vision and Pattern Recognition},
  year         = {2021},
  doi          = {10.1109/CVPR46437.2021.01501},
}

@inproceedings{AdvFlow,
  author       = {Hadi Mohaghegh Dolatabadi and
                  Sarah M. Erfani and
                  Christopher Leckie},
  title        = {AdvFlow: Inconspicuous Black-box Adversarial Attacks using Normalizing
                  Flows},
  booktitle    = {Advances in Neural Information Processing Systems},
  year         = {2020},
}

@inproceedings{CG-attack,
  author       = {Yan Feng and
                  Baoyuan Wu and
                  Yanbo Fan and
                  Li Liu and
                  Zhifeng Li and
                  Shu{-}Tao Xia},
  title        = {Boosting Black-Box Attack with Partially Transferred Conditional Adversarial
                  Distribution},
  booktitle    = {Conference on Computer Vision and Pattern Recognition},
  year         = {2022},
  doi          = {10.1109/CVPR52688.2022.01467},
}

@article{MCG-attack,
  author       = {Fei Yin and
                  Yong Zhang and
                  Baoyuan Wu and
                  Yan Feng and
                  Jingyi Zhang and
                  Yanbo Fan and
                  Yujiu Yang},
  title        = {Generalizable Black-Box Adversarial Attack With Meta Learning},
  journal      = {{IEEE} Transactions on Pattern Analysis and Machine Intelligence},
  volume       = {46},
  number       = {3},
  pages        = {1804--1818},
  year         = {2024},
  doi          = {10.1109/TPAMI.2022.3194988},
}

@inproceedings{difattack,
  author       = {Jun Liu and
                  Jiantao Zhou and
                  Jiandian Zeng and
                  Jinyu Tian},
  title        = {DifAttack: Query-Efficient Black-Box Adversarial Attack via Disentangled
                  Feature Space},
  booktitle    = {AAAI Conference on Artificial Intelligence},
  year         = {2024},
  doi          = {10.1609/AAAI.V38I4.28156},
}

@misc{flux-2-klein,
    author={Black Forest Labs},
    title={FLUX.2: [klein]},
    year={2025},
    howpublished={\url{https://huggingface.co/black-forest-labs/FLUX.2-klein-9b-kv}},
}

@article{gemma,
  author       = {Gemma Team},
  title        = {Gemma: Open Models Based on Gemini Research and Technology},
  journal      = {CoRR},
  volume       = {abs/2403.08295},
  year         = {2024},
  doi          = {10.48550/ARXIV.2403.08295},
}

@inproceedings{CF,
  author       = {Ali Shahin Shamsabadi and
                  Ricardo S{\'{a}}nchez{-}Matilla and
                  Andrea Cavallaro},
  title        = {ColorFool: Semantic Adversarial Colorization},
  booktitle    = {Conference on Computer Vision and Pattern Recognition},
  year         = {2020},
  doi          = {10.1109/CVPR42600.2020.00123},
}

@inproceedings{ddim,
  author       = {Jiaming Song and
                  Chenlin Meng and
                  Stefano Ermon},
  title        = {Denoising Diffusion Implicit Models},
  booktitle    = {International Conference on Learning Representations},
  year         = {2021},
}

@inproceedings{advtex1,
  author       = {Naufal Suryanto and
                  Yongsu Kim and
                  Hyoeun Kang and
                  Harashta Tatimma Larasati and
                  Youngyeo Yun and
                  Thi{-}Thu{-}Huong Le and
                  Hunmin Yang and
                  Se{-}Yoon Oh and
                  Howon Kim},
  title        = {DTA: Physical Camouflage Attacks using Differentiable Transformation
                  Network},
  booktitle    = {Conference on Computer Vision and Pattern Recognition},
  year         = {2022},
  doi          = {10.1109/CVPR52688.2022.01487},
}

@inproceedings{advtex2,
  author       = {Zhanhao Hu and
                  Siyuan Huang and
                  Xiaopei Zhu and
                  Fuchun Sun and
                  Bo Zhang and
                  Xiaolin Hu},
  title        = {Adversarial Texture for Fooling Person Detectors in the Physical World},
  booktitle    = {Conference on Computer Vision and Pattern Recognition},
  year         = {2022},
  doi          = {10.1109/CVPR52688.2022.01295},
}

@article{advbackground,
  author       = {Jiawei Lian and
                  Shaohui Mei and
                  Xiaofei Wang and
                  Yi Wang and
                  Lefan Wang and
                  Yingjie Lu and
                  Mingyang Ma and
                  Lap{-}Pui Chau},
  title        = {Attack Anything: Blind DNNs via Universal Background Adversarial Attack},
  journal      = {CoRR},
  year         = {2024},
  doi          = {10.48550/ARXIV.2409.00029},
}

@inproceedings{advweather1,
  author       = {Alberto Marchisio and
                  Giovanni Caramia and
                  Maurizio Martina and
                  Muhammad Shafique},
  title        = {fakeWeather: Adversarial Attacks for Deep Neural Networks Emulating
                  Weather Conditions on the Camera Lens of Autonomous Systems},
  booktitle    = {International Joint Conference on Neural Networks},
  year         = {2022},
  doi          = {10.1109/IJCNN55064.2022.9892612},
}

@inproceedings{advweather2,
  author       = {Harshitha Machiraju and
                  Vineeth N. Balasubramanian},
  title        = {A Little Fog for a Large Turn},
  booktitle    = {Winter Conference on Applications of Computer Vision},
  year         = {2020},
  doi          = {10.1109/WACV45572.2020.9093549},
}

@inproceedings{imagenet,
  author       = {Alexey Kurakin and
                  Ian J. Goodfellow and
                  Samy Bengio},
  title        = {Adversarial examples in the physical world},
  booktitle    = {5th International Conference on Learning Representations},
  year         = {2017},
}

@inproceedings{resnet,
  author       = {Kaiming He and
                  Xiangyu Zhang and
                  Shaoqing Ren and
                  Jian Sun},
  title        = {Deep Residual Learning for Image Recognition},
  booktitle    = {Conference on Computer Vision and Pattern Recognition},
  year         = {2016},
  doi          = {10.1109/CVPR.2016.90},
}

@inproceedings{convnext,
  author       = {Zhuang Liu and
                  Hanzi Mao and
                  Chao{-}Yuan Wu and
                  Christoph Feichtenhofer and
                  Trevor Darrell and
                  Saining Xie},
  title        = {A ConvNet for the 2020s},
  booktitle    = {Conference on Computer Vision and Pattern Recognition},
  year         = {2022},
  doi          = {10.1109/CVPR52688.2022.01167},
}

@inproceedings{swin,
  author       = {Ze Liu and
                  Yutong Lin and
                  Yue Cao and
                  Han Hu and
                  Yixuan Wei and
                  Zheng Zhang and
                  Stephen Lin and
                  Baining Guo},
  title        = {Swin Transformer: Hierarchical Vision Transformer using Shifted Windows},
  booktitle    = {International Conference on Computer Vision},
  year         = {2021},
  doi          = {10.1109/ICCV48922.2021.00986},
}

@inproceedings{deit,
  author       = {Hugo Touvron and
                  Matthieu Cord and
                  Matthijs Douze and
                  Francisco Massa and
                  Alexandre Sablayrolles and
                  Herv{\'{e}} J{\'{e}}gou},
  title        = {Training data-efficient image transformers {\&} distillation through
                  attention},
  booktitle    = {International Conference on Machine Learning},
  year         = {2021},
}

@inproceedings{vim,
  author       = {Lianghui Zhu and
                  Bencheng Liao and
                  Qian Zhang and
                  Xinlong Wang and
                  Wenyu Liu and
                  Xinggang Wang},
  title        = {Vision Mamba: Efficient Visual Representation Learning with Bidirectional
                  State Space Model},
  booktitle    = {International Conference on Machine Learning},
  year         = {2024},
}

@inproceedings{mambavision,
  author       = {Ali Hatamizadeh and
                  Jan Kautz},
  title        = {MambaVision: {A} Hybrid Mamba-Transformer Vision Backbone},
  booktitle    = {Conference on Computer Vision and Pattern Recognition},
  year         = {2025},
  doi          = {10.1109/CVPR52734.2025.02352},
}

@inproceedings{robust_models,
  author       = {Florian Tram{\`{e}}r and
                  Alexey Kurakin and
                  Nicolas Papernot and
                  Ian J. Goodfellow and
                  Dan Boneh and
                  Patrick D. McDaniel},
  title        = {Ensemble Adversarial Training: Attacks and Defenses},
  booktitle    = {International Conference on Learning Representations},
  year         = {2018},
}

@inproceedings{vit,
  author       = {Alexey Dosovitskiy and
                  Lucas Beyer and
                  Alexander Kolesnikov and
                  Dirk Weissenborn and
                  Xiaohua Zhai and
                  Thomas Unterthiner and
                  Mostafa Dehghani and
                  Matthias Minderer and
                  Georg Heigold and
                  Sylvain Gelly and
                  Jakob Uszkoreit and
                  Neil Houlsby},
  title        = {An Image is Worth 16x16 Words: Transformers for Image Recognition
                  at Scale},
  booktitle    = {International Conference on Learning Representations},
  year         = {2021},
}

@inproceedings{inception,
  author       = {Christian Szegedy and
                  Vincent Vanhoucke and
                  Sergey Ioffe and
                  Jonathon Shlens and
                  Zbigniew Wojna},
  title        = {Rethinking the Inception Architecture for Computer Vision},
  booktitle    = {Conference on Computer Vision and Pattern Recognition},
  year         = {2016},
  doi          = {10.1109/CVPR.2016.308},
}

@article{NIMA,
  title={NIMA: Neural Image Assessment},
  author={Hossein Talebi and Peyman Milanfar},
  journal={IEEE Transactions on Image Processing},
  year={2017},
  volume={27},
  pages={3998-4011},
}

@inproceedings{hyperiqa,
  author       = {Shaolin Su and
                  Qingsen Yan and
                  Yu Zhu and
                  Cheng Zhang and
                  Xin Ge and
                  Jinqiu Sun and
                  Yanning Zhang},
  title        = {Blindly Assess Image Quality in the Wild Guided by a Self-Adaptive
                  Hyper Network},
  booktitle    = {Conference on Computer Vision and Pattern Recognition},
  year         = {2020},
  doi          = {10.1109/CVPR42600.2020.00372},
}

@inproceedings{musiq,
  title={MUSIQ: Multi-scale Image Quality Transformer},
  author={Junjie Ke and Qifei Wang and Yilin Wang and Peyman Milanfar and Feng Yang},
  booktitle={International Conference on Computer Vision},
  year={2021},
}

@inproceedings{tres,
  title={No-Reference Image Quality Assessment via Transformers, Relative Ranking, and Self-Consistency},
  author={S. Alireza Golestaneh and Saba Dadsetan and Kris M. Kitani},
  booktitle={Winter Conference on Applications of Computer Vision},
  year={2021},
}

@inproceedings{imagenet_val,
  title={ImageNet: A large-scale hierarchical image database},
  author={Jia Deng and Wei Dong and Richard Socher and Li-Jia Li and K. Li and Li Fei-Fei},
  booktitle={Conference on Computer Vision and Pattern Recognition},
  year={2009},
}

@article{dinov2,
  author       = {Maxime Oquab and
                  Timoth{\'{e}}e Darcet and
                  Th{\'{e}}o Moutakanni and
                  Huy V. Vo and
                  Marc Szafraniec and
                  Vasil Khalidov and
                  Pierre Fernandez and
                  Daniel Haziza and
                  Francisco Massa and
                  Alaaeldin El{-}Nouby and
                  Mido Assran and
                  Nicolas Ballas and
                  Wojciech Galuba and
                  Russell Howes and
                  Po{-}Yao Huang and
                  Shang{-}Wen Li and
                  Ishan Misra and
                  Michael Rabbat and
                  Vasu Sharma and
                  Gabriel Synnaeve and
                  Hu Xu and
                  Herv{\'{e}} J{\'{e}}gou and
                  Julien Mairal and
                  Patrick Labatut and
                  Armand Joulin and
                  Piotr Bojanowski},
  title        = {DINOv2: Learning Robust Visual Features without Supervision},
  journal      = {Transactions on Machine Learning Research},
  year         = {2024},
}

@inproceedings{stable-flow,
  author       = {Omri Avrahami and
                  Or Patashnik and
                  Ohad Fried and
                  Egor Nemchinov and
                  Kfir Aberman and
                  Dani Lischinski and
                  Daniel Cohen{-}Or},
  title        = {Stable Flow: Vital Layers for Training-Free Image Editing},
  booktitle    = {Conference on Computer Vision and Pattern Recognition},
  year         = {2025},
  doi          = {10.1109/CVPR52734.2025.00738},
}

@inproceedings{wrn,
  author       = {Sergey Zagoruyko and
                  Nikos Komodakis},
  title        = {Wide Residual Networks},
  booktitle    = {British Machine Vision Conference},
  year         = {2016},
}

@Article{fid,
  author      = {Fr{\'e}chet, Maurice},
  journal     = {{Annales de l'ISUP}},
  title       = {{Sur la distance de deux lois de probabilit{\'e}}},
  year        = {1957},
  number      = {3},
  pages       = {183-198},
  volume      = {VI},
  hal_version = {v1},
}

\end{document}